\documentclass{article}

\usepackage[nonatbib,final]{nips_2017}

\usepackage[utf8]{inputenc} 
\usepackage[T1]{fontenc}    
\usepackage{hyperref}       
\usepackage{url}            
\usepackage{booktabs}       
\usepackage{amsfonts}       
\usepackage{nicefrac}       
\usepackage{microtype}      
\usepackage{graphicx}

\title{Mental Sampling in Multimodal Representations}

\author{
  Jian-Qiao~Zhu \\
  Department of Psychology\\
  University of Warwick\\
  \texttt{j.zhu@warwick.ac.uk} \\
  \And
  Adam N.~Sanborn\\
  Department of Psychology\\
  University of Warwick\\
  \texttt{a.n.sanborn@warwick.ac.uk} \\
  \AND
  Nick Chater\\
  Behavioural Science Group\\
  Warwick Business School\\
  \texttt{nick.chater@wbs.ac.uk} \\
}

\begin{document}

\maketitle

\begin{abstract}
Both resources in the natural environment and concepts in a semantic space are distributed ``patchily'', with large gaps in between the patches. To describe people's internal and external foraging behavior, various random walk models have been proposed. In particular, internal foraging has been modeled as sampling: in order to gather relevant information for making a decision, people draw samples from a mental representation using random-walk algorithms such as Markov chain Monte Carlo (MCMC). However, two common empirical observations argue against simple sampling algorithms such as MCMC. First, the spatial structure is often best described by a L\'{e}vy flight distribution: the probability of the distance between two successive locations follows a power-law on the distances. Second, the temporal structure of the sampling that humans and other animals produce have long-range, slowly decaying serial correlations characterized as $1/f$-like fluctuations. We propose that mental sampling is not done by simple MCMC, but is instead adapted to multimodal representations and is implemented by Metropolis-coupled Markov chain Monte Carlo (MC$^3$), one of the first algorithms developed for sampling from multimodal distributions. MC$^3$ involves running multiple Markov chains in parallel but with target distributions of different temperatures, and it swaps the states of the chains whenever a better location is found. Heated chains more readily traverse valleys in the probability landscape to propose moves to far-away peaks, while the colder chains make the local steps that explore the current peak or patch. We show that MC$^3$ generates distances between successive samples that follow a L\'{e}vy flight distribution and $1/f$-like serial correlations, providing a single mechanistic account of these two puzzling empirical phenomena.
\end{abstract}

\section{Introduction}

In many complex domains, such as vision, motor control, language, categorization or common-sense reasoning, human behavior is consistent with the predictions of Bayesian models (e.g., \cite{battaglia2013, sanborn2013, chater2006, anderson1991, griffiths2011, kemp2009, wolpert2007, yuille2006}). Bayes' theorem prescribes a simple normative method to combine prior beliefs with new information to make inferences about the world. However, the sheer number of hypotheses that must be considered in complex domains makes exact Bayesian inference intractable. Instead it must be that individuals are performing some kind of approximate inference \cite{vul2014, sanborn2016}. 

Sampling is a way to perform approximation for Bayesian models in complex problems that makes many difficult computations easy: instead of integrating over vast hypothesis spaces, samples of hypotheses can be drawn from the posterior distribution. The computational cost of sample-based approximations only scales with the number of samples rather than the dimensionality of the hypothesis space, though using small numbers of samples result in particular biases in inference. 

Interestingly, the biases in inference that are introduced by using a small number of samples match some of the biases observed in human behavior. For example, probability matching \cite{vul2014}, anchoring effects \cite{lieder2012}, and many reasoning fallacies \cite{dasgupta2016, sanborn2016} can all be explained in this way. However, there is as of yet no consensus on the exact nature of the algorithm used to sample from human mental representations.

Previous work has posited that people either use direct sampling or Markov chain Monte Carlo (MCMC) to sample from their posterior distribution over hypotheses \cite{vul2014,lieder2012,dasgupta2016,sanborn2016}. In this paper, we demonstrate that these algorithms cannot explain two key empirical effects that have been found in a wide variety of tasks. In particular, these algorithms do not produce distances between samples that follow a L\'{e}vy flight distribution, and separately they do not produce autocorrelations that follow $1/f$ scaling. To find a sampling algorithm that does match these empirical effects, we note that mental representations have been shown to be ``patchy'' with high probability regions separated by large regions of low probability. We then compare one of the first algorithms developed for sampling from multimodal distributions, Metropolis-coupled MCMC (MC$^3$), and demonstrates that it produces both key empirical phenomena. Previously L\'{e}vy flight distributions and $1/f$ scaling have been explained separately as the result of efficient search and the signal of self-organizing behavior respectively \cite{viswanathan1999,vanorden2003}, and we provide the first account that can explain both phenomena as the result of the same purposeful mental activity.

\subsection{Spatial structure of mental samples}

In the real world, resources are rarely distributed uniformly in the environment. Food, water, and other critical nature resources often occur in spatially isolated patches with large gaps in between. Therefore, humans and other animals' foraging behaviors should adapt to such patchy environments. In fact, foraging behaviors have been observed to display a L\'{e}vy flight, which is a class of random walk whose step lengths follow a heavy-tailed power-law distribution \cite{shlesinger1995}. In a L\'{e}vy flight distribution, the probability of executing a jump of length $l$ is given by:

\begin{equation}
P(l) \sim l^{-\mu}
\end{equation}

where $1<\mu\leq 3$. The values $\mu\leq 1$ do not correspond to normalizable probability distributions. Examples of mobility patterns following the L\'{e}vy flight has been recorded in Albatrosses \cite{viswanathan1996}, marine predators \cite{sims2008}, monkeys \cite{ramos2004}, and humans \cite{gonzalez2008}. 

L\'{e}vy flights are advantageous in patchy environments where resources are sparsely and randomly distributed because the probability of returning to a previously visited target site is smaller than in a standard random walk. In the same patchy environment, L\'{e}vy flights can visit more new target sites than a random walk does \cite{berkolaiko1996}. Interestingly, it has been proven that in foraging the optimal exponent is $\mu=2$ regardless the dimensionality of the space if (1) the target sites are sparse, (2) they can be visited any number of times, and (3) the forager can only detect and remember a target site in a close vicinity \cite{viswanathan1999}. 

Remarkably, mental representations of concepts are also patchy and the distance between mental samples also follows a L\'{e}vy flight distribution. For example, in semantic fluency tasks (e.g., asking participants to ``name as many animals as you can''), the retrieved animals tend to form clusters (e.g., pets, water animals, African animals) \cite{troyer1997, abbott2012}. This same task has also been found to produce L\'{e}vy flight distributions of inter-response intervals (IRI) \cite{rhodes2007}, which can be considered a measure of distance between samples by making the reasonable assumption that there is a linear relationship between IRI and mental distance\footnote{There are various ways to make the link between IRI and distance between samples. One is to assume that it takes longer to transition to a sample that is further away in the mental space. A second is to assume that while generating any sample takes the same fixed amount of time, there are unreported samples that are generated between each reported sample, and that the sampler has travelled further the more unreported samples that are generated; unreported samples are plausible in this task because participants are only given credit for each new animal name they report.}.

\subsection{Temporal structure of mental samples}

Besides the spatial structure of the distance between two successive locations following a power-law distribution, a number of studies has reported that the temporal structure of many cognitive activities contains long-range, slowly decaying serial correlations. These correlations tend to follow a $1/f$ scaling law \cite{kello2010}:
\begin{equation}
C(k) \sim k^{-\alpha}
\end{equation}
where $C(k)$ is the autocorrelation function of temporal lag $k$. The same phenomenon is often expressed in the frequency domain:
\begin{equation}
S(f) \sim f^{-\alpha}
\end{equation}
where $f$ is frequency, $S(f)$ is spectral power and $\alpha \in [0.5, 1.5]$ is considered $1/f$ scaling. The power spectra can be derived from submitting the time series to Fourier analysis. $1/f$ noise is also known as pink or flicker noise, which varies in predictability intermediately between white noise (no serial correlation,  $S(f) \sim 1/f^0$) and brown noise (no correlation between increments,  $ S(f) \sim 1/f^2$). Note that L\'{e}vy flights are random walks so they do not produce $1/f$ noise, but $1/f^2$ noise instead. 

$1/f$-like temporal fluctuations in human cognition were first reported in time estimation and spatial interval estimation tasks in which participants were asked to repeatedly estimate a pre-determined time interval of 1 second or spatial interval of 1 inch \cite{gilden1995}. Subsequent studies have shown $1/f$ scaling laws in the response times of mental rotation, lexical decision, serial visual search, and parallel visual search \cite{gilden1997}, as well as the time to switch between different percepts when looking at a bistable stimulus (i.e., a Necker cube \cite{gao2006}).

Given that sampling can be described as a L\'{e}vy flight spatially and has $1/f$ autocorrelations (see Table~\ref{summary} for summary), we now investigate which sampling algorithms can capture both the spatial and temporal structure of human cognition.

\begin{center}
\begin{table}[h]
  \caption{Key findings for $1/f$ noise and L\'{e}vy flight in human mental sampling}
  \label{summary}
  \centering
  \begin{tabular}{lll}
    \toprule
     Papers     & Experiments     & Main findings \\
    \midrule
    \cite{gilden1995}  & Time interval estimation & Power spectra slopes of $[-1.2,-0.90]$ \\
    & Spatial interval estimation & Power spectra slope of $-1$ \\
    \cite{gilden1997} & Mental rotation & RT power spectra slope of $-0.7$ \\
    & Lexical decision & RT power spectra slope of $-0.9$ \\
    & Serial search & RT power spectra slope of $-0.7$ \\
    & Parallel search & RT power spectra slope of $-0.7$ \\
    \cite{rhodes2007} & Memory retrieval task & IRI power-law exponents $\hat{\mu}\in[1.37,1.98]$ \\
    \cite{rhodes2011} & Natural scene perception & Eye movement trajectories follow both $1/f$ noise and L\'{e}vy flight \\
    \bottomrule
  \end{tabular}
\end{table}
\end{center}

\section{Sampling algorithms}
We consider three possible sampling algorithms that might be employed in human cognition: Direct Sampling (DS), Random walk Metropolis (RwM), and Metropolis-coupled MCMC (MC$^3$). We define DS as independently drawing samples in accord with the posterior probability distribution. DS is the most efficient algorithm for sampling of the three, but it may not be possible to implement in human cognition as it often requires calculating intractable normalizing constants that scale exponentially with the dimensionality of the hypothesis space \cite{mackay2003, chater2006a}. DS has been used to explain biases in human cognition such as probability matching \cite{vul2014}.

MCMC algorithms can bypass the problem of the normalizing constant by simulating a Markov chain that transitions between states according only to the ratio of the probability of hypotheses \cite{mackay2003}. We define RwM as a classical Metropolis-Hastings MCMC algorithm, which can be thought of as a random walker exploring the probability landscape of hypotheses, preferentially climbing the peaks of the posterior probability distribution \cite{metropolis1953, hastings1970}. However, with limited number of samples, RwM is very unlikely to reach modes in the probability distribution that are separated by large regions of low probability. This leads to biased approximations of the posterior distribution \cite{swendsen1986, sanborn2016}. Random walks have been used to model clustered responses in memory retrieval \cite{abbott2012}, and RwM in particular has been used to model multistable perception \cite{gershman2012}, the anchoring effect \cite{lieder2012}, and various reasoning biases \cite{dasgupta2016, sanborn2016}.

Our third algorithm is MC$^3$, also known as parallel tempering or replica-exchange MCMC, was one of the first algorithms to successfully tackle the problem of multimodality \cite{geyer1991}. MC$^3$ involves running $M$ Markov chains in parallel, each at a different temperature: $T_1, T_2, ... ,T_M$. In general, $1=T_1<T_2< ... <T_M$, and $T_1$ is the temperature of the interest where the target distribution is unchanged. The purpose of the heated chains is to traverse valleys in the probability landscape to propose moves to far-away peaks (by sampling from heated target distributions: $\pi^{1/T}$), while the colder chains make the local steps that explore the current probability peak or patch. MC$^3$ decides whether to swap the states between two randomly chosen chains in every iteration \cite{geyer1991}. In particular, swapping of chain $i$ and $j$ is accepted or rejected according to a Metropolis rule; hence, the name Metropolis-coupled MCMC

\begin{equation}
A^{swap} = \min \{1, \frac{\pi(x_j)^{1/T_i}  \pi(x_i)^{1/T_j} }{\pi(x_i)^{1/T_i} \pi(x_j)^{1/T_j}}\}
\end{equation}

Coupling induces dependence among the chains, so each chain is no longer Markovian. The stationary distribution of the entire set of chains is thus $\prod_{i=1}^M \pi^{1/T_i}$ but we only use samples from the cold chain ($T=1$) to approximate the posterior distribution \cite{geyer1991}. Pseudocode for MC$^3$ is presented below. Note that MC$^3$ reduces to RwM when the number of parallel chains $M=1$.

\begin{center}
\begin{tabular}{rlr}
\hline
  \multicolumn{2}{l}{\textbf{Algorithm} Metropolis-coupled Markov chain Monte Carlo} \\ \hline
 1: & Choose a starting point $x_1$. \\
 2: & \textbf{for} $t=2$ to $L$ \textbf{do} \\
 3: & ~~~~ \textbf{for} $m=1$ to $M$ \textbf{do} & $\rhd$ update all $M$ chains \\
 4: & ~~~~~~~~ Draw a candidate sample $x' \sim \mathcal{N}(x_{t-1}^m, \sigma)$  & $\rhd$ Gaussian proposal distribution \\
 5: & ~~~~~~~~ Sample $u \sim U[0,1]$ \\
 6: & ~~~~~~~~ $A^m= \min \{ 1, [\frac{\pi(x')}{\pi(x_{t-1}^m)}]^{1/T_m}\}$    \\
 7: & ~~~~~~~~ \textbf{if} $u<A^m$ \textbf{then} $x_t^m = x'$  \textbf{else} $x_t^m=x_{t-1}^m$ \textbf{end if} & $\rhd$ Metropolis acceptance rule \\
 8: & ~~~~ \textbf{end for} \\
 9: & ~~~~ \textbf{repeat} floor($M/2$) \textbf{times} & $\rhd$  swapping scheme for Markov chains \\
 10: & ~~~~~~~~ Randomly select two chain $i,j$ without repetition  \\
 11: & ~~~~~~~~ Sample $u \sim U[0,1]$ \\
 12: & ~~~~~~~~ $A^{swap} = \min \{1, \frac{\pi(x_t^j)^{1/T_i}  \pi(x_t^i)^{1/T_j} }{\pi(x_t^i)^{1/T_i} \pi(x_t^j)^{1/T_j}}\}$  \\
 13: & ~~~~~~~~ \textbf{if} $u<A^{swap}$ \textbf{then} swap($x_t^i, x_t^j$) \textbf{end if} & $\rhd$ Metropolis-coupled swapping rule \\
 14: & ~~~~ \textbf{end repeat} \\
 15: & \textbf{end for} \\
\hline  
\end{tabular}
\end{center}

\section{Results}

In this section, we evaluate whether the two key empirical effects of L\'{e}vy flights and $1/f$ autocorrelations can be produced by the Direct Sampling (DS), Random walk Metropolis (RwM), and Metropolis-coupled MCMC (MC$^3$) algorithms.

\subsection{L\'{e}vy flight}

We simulated a 2D patchy environment with $N_{mode}=15$ Gaussian mixtures where the means are uniformly generated from $[-r, r]$ for both dimensions, where $r=9$ and the covariance matrix is fixed as the identity matrix for all mixtures. This method will produce a patchy environment (for example the top panel of Figure 1). We ran DS, RwM, and MC$^3$ on this multimodal probability landscape, and the first 100 positions for each algorithm can be found in the top panel of Figure 1. The empirical flight distances were obtained by calculating the Euclidean distance between two consecutive positions of the sampler. For MC$^3$, only the positions of the cold chain ($T=1$) were used.

Power-law distributions should produce straight lines in a log-log plot. Therefore, the power-law exponents were fitted by linear regression on the window-averaged log-binned flight distance data \cite{rhodes2007}. We used 10 non-overlapping windows that evenly split the x-axis, and cell means are represented in the yellow filled dots in the bottom panel of Figure 1. Fitting the cell means provides a lower-variance method for estimating the slope than fitting the log-binned data directly. Figure 1 (bottom panel) shows that only MC$^3$ can reproduce the distributional property of flight distance as a L\'{e}vy flight with estimated power-law exponent $\hat{\mu}=1.27$. Both DS ($\hat{\mu}=-0.59$) and RwM ($\hat{\mu}=0.91$) \footnote{The log-binned data for RwM shows that the algorithm is certainly not producing a power-law as the empirical flight distance distribution is not a straight line in the log-log plot} produced values outside the range of a L\'{e}vy flight. 

\begin{figure}[h]
  \centering
  \includegraphics[width=\textwidth]{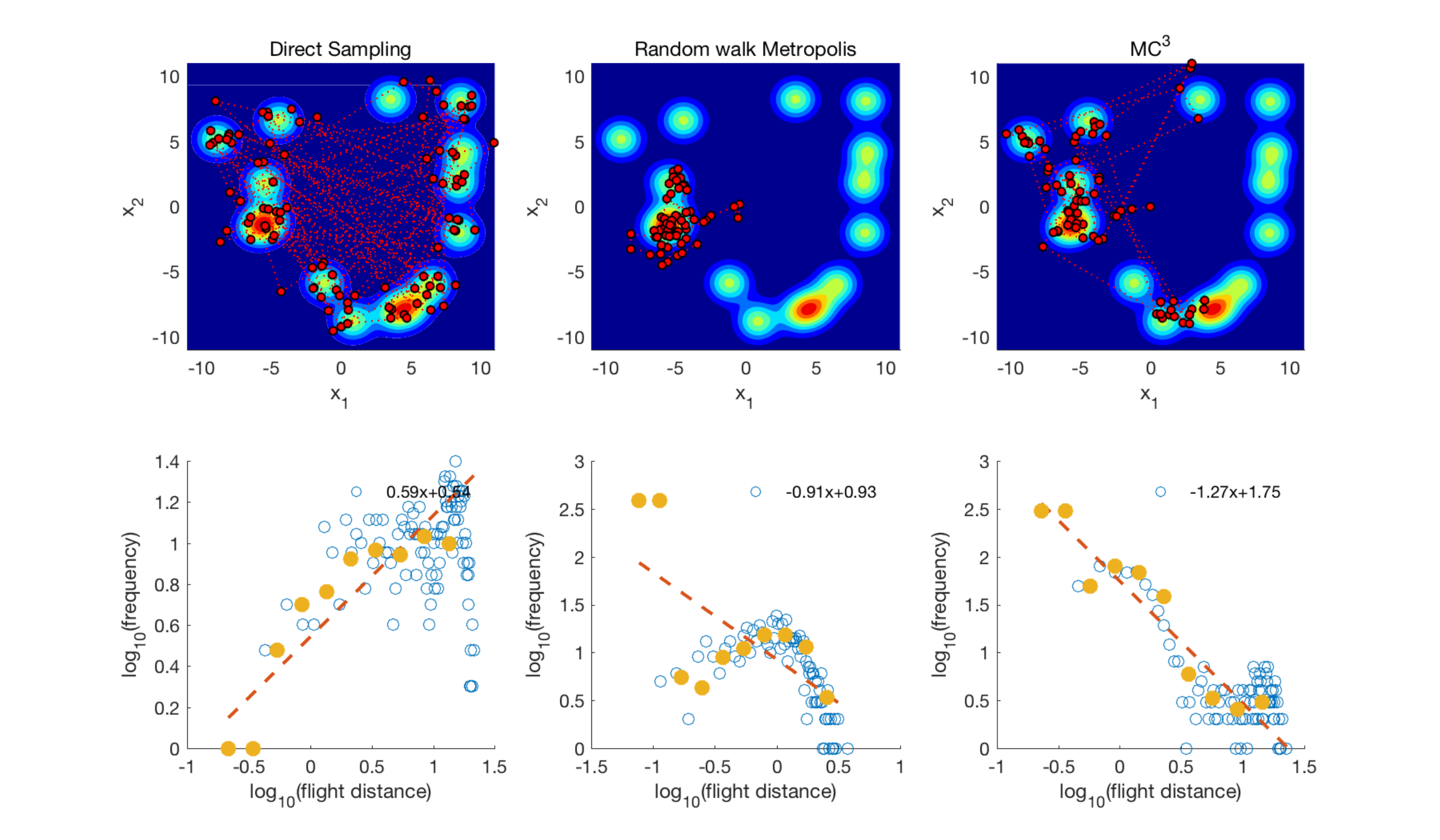}
  \caption{Searching behavior in a simulated 2D patchy environment of a 15 component Gaussian mixture. (\textit{Left Panel}) the trajectory of first 100 positions (red dots in top panel) and the log-log plot of flight distance (bottom panel) for DS. The best-fitted lines used to estimate the Levy flight exponent (bottom panel red dashed lines) were based on cell means using non-overlapping windows (yellow filled dots) of the log-binned data (blue dots). (\textit{Middle Panel}) the same plots for RwM algorithm. The Gaussian proposal distribution was an identity covariance matrix. (\textit{Right Panel}) the same plots for MC$^3$ algorithm with 8 parallel chains with only the cold chain shown here. The Gaussian proposal distributions for all 8 chains had the same identity covariance matrix. For all the algorithms only the first 1024 samples were used.}
\end{figure}

We then investigated the impact of spatial sparsity on the estimated power-law exponents. In this simulation, the same number of Gaussian mixture were used but the range $r$ was varied. The spatial sparsity was computed as the mean distance between Gaussian modes. With small or moderate spatial sparsity we found a positive relationship between spatial sparsity and the estimated power-law exponents for both DS and MC$^3$ (see Figure 2 left). In this range, only MC$^3$ produced power-law exponents in the range reported in human mental foraging studies (see Table 1), while both DS and RwM failed to do so. For all three algorithms, once spatial sparsity was too great only a single mode was explored and no large jumps were made.

We also checked whether MC$^3$ really is more suitable to explore patchy mental representations than RwM. In our simulated patchy environment, which used 15 identical Gaussian mixtures with identity covariance matrix, an optimal sampling algorithm should visit each mode equally often, hence will produce a uniform distribution of visit frequencies over all the modes. To this end, the effectiveness of exploring the representation was examined by computing a Kullback-Leibler divergence (KL) \cite{mackay2003} between a uniform distribution over all modes and a the relative frequency of how often an algorithm visited each mode:

\begin{equation}
D_{KL} (\mathcal{H}_{1:t} || U) = \sum_{i=1}^{N_{mode}} \mathcal{H}_{1:t} \log \frac{\mathcal{H}_{1:t}}{1/N_{mode}}
\end{equation}

where $U$ is a discrete uniform distribution, $N_{mode}$ is the number of identical Gaussian mixtures, and $\mathcal{H}$ is the empirical frequency of visited modes up to time $t$. Samples were assigned to the closest mode when determining these empirical frequencies. The faster the KL divergence for an algorithm reaches zero, the more effective the algorithm is at exploring the underlying environment and the DS algorithm serves as a benchmark for the other two algorithms.  As shown in Figure 2 (middle), MC$^3$ quickly catches up to DS, while RwM lags far behind in exploring this patchy environment.

\begin{figure}[h]
  \centering
  \includegraphics[width=\textwidth]{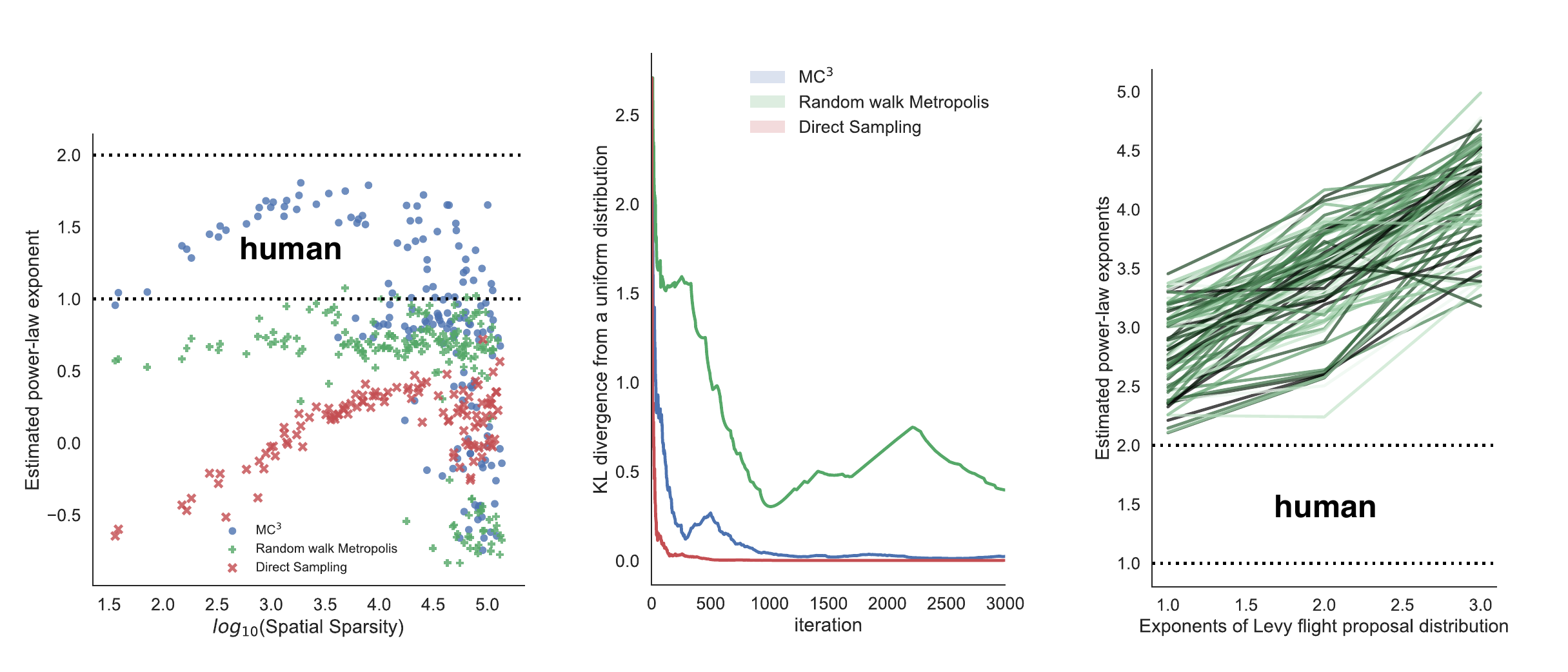}
  \caption{(\textit{Left}) Estimated power-law exponents for flight distance distributions for the three sampling algorithms, manipulating the spatial sparsity of the Gaussian mixture environment. Spatial sparsity measurement was defined as the mean distance between modes. All three algorithms used the same settings as in Figure 1. The dashed lines show the range of human data. (\textit{Middle}) KL divergence of mode visiting from the true distribution for the three sampling algorithms. The underlying patchy environments are the same for all three algorithms. (\textit{Right}) Simulated RwM with a L\'{e}vy flight proposal distribution. Darker colors represent higher spatial sparsities. The dashed lines show the range of human data.}
\end{figure}

Of course it may seem that we were simply using the wrong proposal distribution for RwM. Instead of using a Gaussian proposal distribution we can use a L\'{e}vy flight proposal distribution, which will straightforwardly produce L\'{e}vy flights if the posterior distribution is uniform over the entire space (i.e., every proposed flight will be accepted). However, in a patchy environment a L\'{e}vy flight proposal distribution will not typically produce a L\'{e}vy flight distribution of distances between samples that has estimated power-law exponents in the range of human data, as also can be seen in Figure 2 (right) with different spatial sparsities. The reason for this is that the long jumps in the proposal distribution are unlikely to be successful: these long jumps often propose new states that lie in regions of nearly zero posterior probability.

\subsection{$1/f$ noise}
A typical interval estimation task requests participants to repeatedly produce an estimation for the same target interval over many repeated trials \cite{gilden1995, gilden1997}. For instance, participants were first given an example of a target interval (e.g., 1 second time interval or 1-inch spatial interval) and then repeated the judgments again and again without feedback for up to 1000 trials. These time series produced by human subjects showed $1/f$ noise, with an exponent close to 1. However, the log-log plot in human data is typically observed flatten out for the highest frequencies \cite{gilden1995}. This effect has been explained as the result of two processes: fractional Brownian motion combined with white noise at the highest frequencies \cite{gilden1995}. 

Figure 3 shows an example of time series for the first 1024 samples generated by DS (left), RwM (middle), and MC$^3$ (right). We used a simple Gaussian target distribution in this simulation because the distribution of responses produced by participants was indistinguishable from a Gaussian \cite{gilden1995}. Note that RwM and MC$^3$ were initiated at the mode of the Gaussian distribution, and there was no burn-in period in our simulation. This results show that only MC$^3$ produces $1/f$ noise ($\hat{\alpha}=1.01$), whereas DS produces white noise ($\hat{\alpha}=-0.01$) and RwM is closest to brown noise ($\hat{\alpha}=1.64$). RwM tends to generate brown noise because, if every proposed sample is accepted, then the algorithm reduces to first-order autoregressive process (i.e., AR(1)) \cite{xu2009}. This is shown through simulation in Figure 4: when the Gaussian width ($\sigma_{target}$) becomes much greater width of the Gaussian proposal distribution ($\sigma_{proposal}$), RwM produces brown noise.

\begin{figure}[h]
  \centering
  \includegraphics[width=0.8\textwidth]{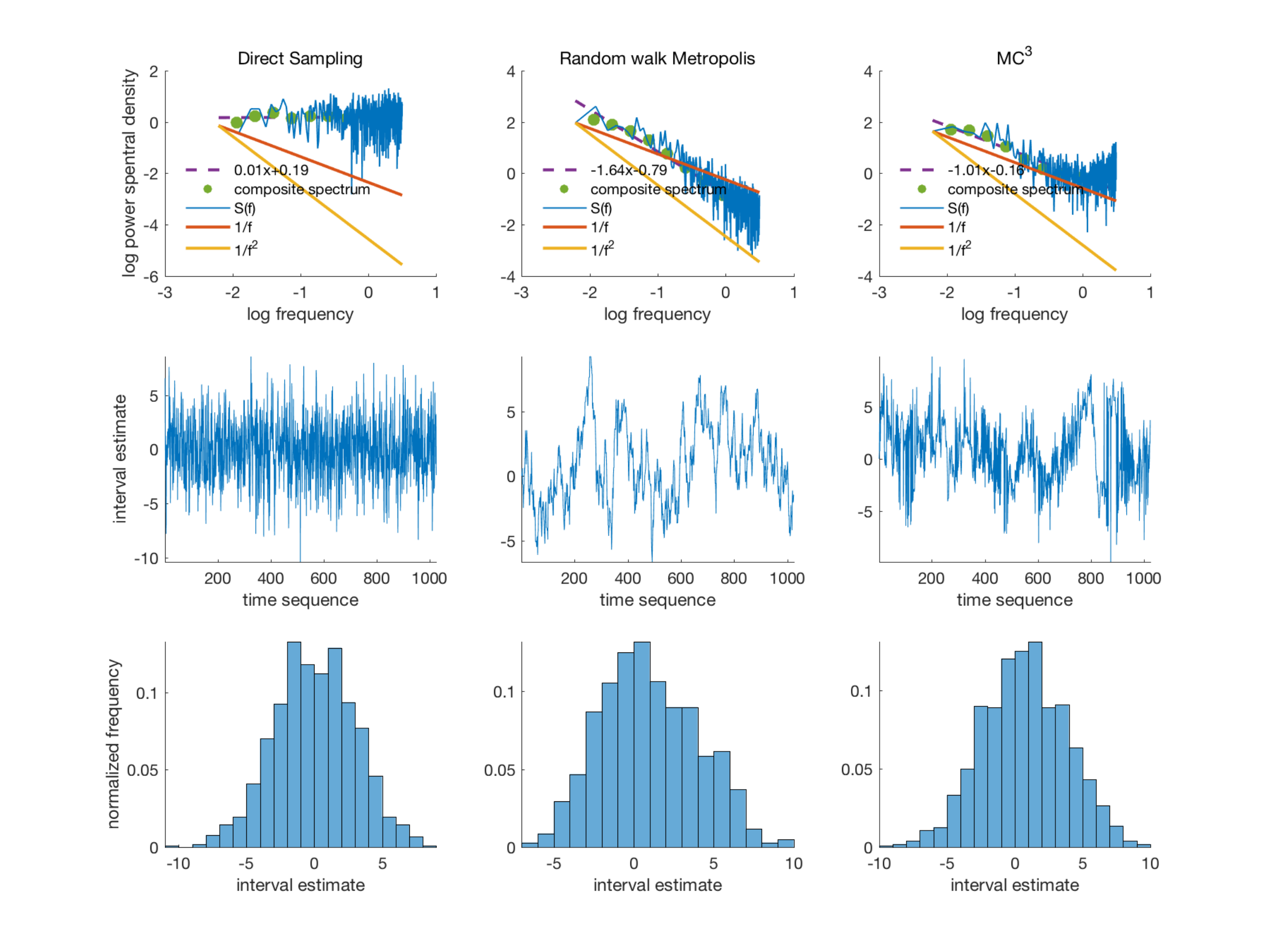}
  \caption{Sampling from a unimodal Gaussian $\mathcal{N}(0,3)$. (\textit{Left Panel}) the power spectra, traceplot, and sample distribution of DS from top to bottom. The best-fitted lines in power spectra (top panel purple dashed lines) are estimated based on block-averaged periodograms (green filled dots; \cite{thornton2005}). (\textit{Middle Panel}) the same plots for RwM algorithm. (\textit{Right Panel}) the same plots for MC$^3$ algorithm with 2 parallel chains and only the cold chain is shown here. The result will be similar to using 8 parallel chains if we restrict swapping between neighboring chains only. For all the algorithms here, the first 1024 samples were used.}
\end{figure}

In contrast, MC$^3$ has a tendency to produce $1/f$ noise when the acceptance rate is high. It has been shown that the sum of as few as three AR(1) processes with widely distributed autoregressive coefficients produces an approximation to $1/f$ noise \cite{ward2002}. As the higher-temperature chains can be thought of as very roughly similar to AR(1) processes with lower autoregressive coefficients, this may explain why the asymptotic behavior of the MC$^3$ is $1/f$ noise.

What is also interesting about MC$^3$ is that it is a single process that is able to produce both the $1/f$ slope at lower frequencies as well as the flattening of the slope at higher frequencies, which was ascribed to two different processes by \cite{gilden1995}. The reason MC$^3$ produces this result appears to be because when two chains with similar temperatures find states with similar posterior probability they will repeatedly swap back and forth, which can produce high frequency oscillations in the coldest chain.

\begin{figure}[h]
  \centering
  \includegraphics[width=0.6\textwidth]{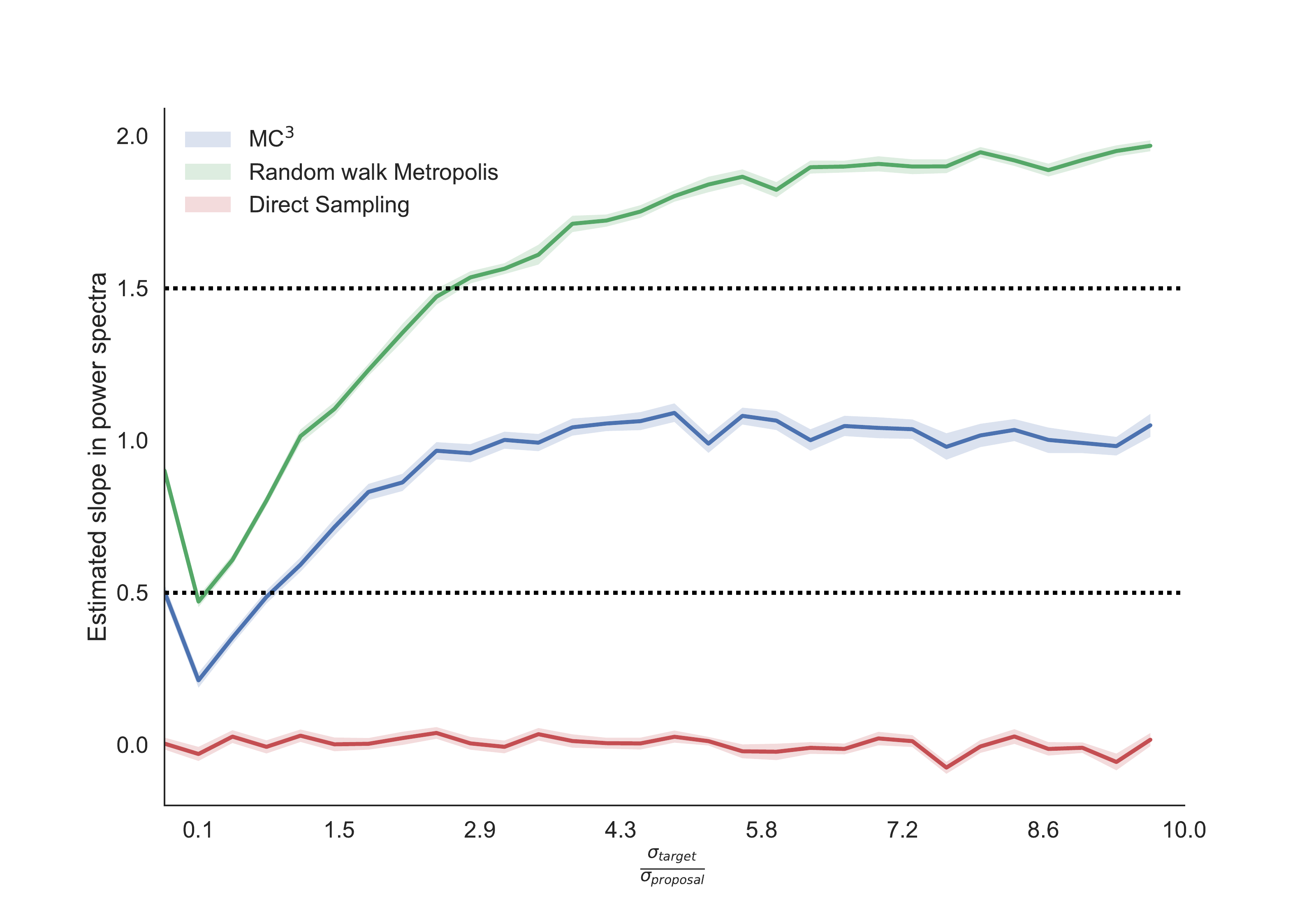}
  \caption{Estimated slopes in the power spectra are related to the ratio between Gaussian width and proposal step size. When the ratio is low, the acceptance rate of proposed sample should be low; it is the opposite case for the high ratio. The asymptotic behaviors of MC$^3$ are $1/f$ noise, of RwM are brown noise, and of DS are white noise.
}
\end{figure}

\section{Discussion}
L\'{e}vy flights are advantageous in a patchy world, and have been observed in many foraging task with humans and other animals. A random walk with Gaussian steps does not produce the occasional long-distance jump as a L\'{e}vy flight does. However, the swapping scheme between parallel chains of MC$^3$ enables it to produce L\'{e}vy-like scaling in the flight distance distribution. Additionally MC$^3$ produces the long-range slowly-decaying temporal correlations of $1/f$ scaling. This long-range dependence rules out any sampling algorithm that draws independent samples from the posterior distribution, such as DS, since the sample sequence would have no serial correlation (i.e., white noise). It also rules out RwM because the current sample solely depends on the previous sample. Both of these results suggest that the algorithms people use to sample mental representations are more complex than DS or RwM, and, like MC$^3$, are instead adapted to sampling from multimodal distributions.

However, if people are adapted to multimodal distributions, their behavior appears to have the same temporal pattern even when they are actually sampling from a unimodal distribution. In Gilden's experiments, the overall distribution of estimated intervals (i.e., ignoring serial order) was not multimodal, instead it was indistinguishable from a Gaussian distribution \cite{gilden1995}. Assuming that the posterior distribution in the hypothesis space is also unimodal then it is somewhat inefficient to use MC$^3$ rather than simple MCMC. Potentially the brain is hardwired to use particular algorithms, or it is slow to adapt to unimodal representations because it is very difficult to know that a distribution is unimodal rather than just a single mode in a patchy space.

Previous explanations of scale-free phenomenon in human cognitions such as self-organized criticality argue that $1/f$ noise is generated from the interactions of many simple processes that produce such hallmarks of complexity \cite{vanorden2003}. Other explanations assume that it is due to a mixture of scaled processes like noise in attention or noise in our ability to perform cognitive tasks \cite{wagenmakers2004}. These approaches argue that $1/f$ noise is a general property of cognition, and do not tie it to other empirical effects. Our explanation of this scale-free process is more mechanistic, assuming that they reflect the cognitive need to gather vital resources in a multimodal world. While autocorrelations make samplers less effective when sampling from simple distributions, they may need to be tolerated in a multimodal world in order to sample other isolated modes.

Of course, we do not claim that MC$^3$ is the only sampling algorithm that is able to produce both $1/f$ noise and L\'{e}vy flights. It is possible that other algorithms that deal better with multimodality than MCMC, such as running multiple non-random walk Markov chains in parallel or Hamiltonian Monte Carlo, could produce similar results. Future work will explore which algorithms can match these key human data.

\bibliographystyle{abbrv}
\bibliography{references}

\begin{thebibliography}{10}

\bibitem{abbott2012}
J.~T. Abbott, J.~L. Austerweil, and T.~L. Griffiths.
\newblock Human memory search as a random walk in a semantic network.
\newblock In {\em Advances in Neural Information Processing Systems}, pages
  3050--3058, 2012.

\bibitem{anderson1991}
J.~R. Anderson.
\newblock The adaptive nature of human categorization.
\newblock {\em Psychological Review}, 98(3):409, 1991.

\bibitem{battaglia2013}
P.~W. Battaglia, J.~B. Hamrick, and J.~B. Tenenbaum.
\newblock Simulation as an engine of physical scene understanding.
\newblock {\em Proceedings of the National Academy of Sciences},
  110(45):18327--18332, 2013.

\bibitem{berkolaiko1996}
G.~Berkolaiko, S.~Havlin, H.~Larralde, and G.~Weiss.
\newblock Expected number of distinct sites visited by n l{\'e}vy flights on a
  one-dimensional lattice.
\newblock {\em Physical Review E}, 53(6):5774, 1996.

\bibitem{chater2006}
N.~Chater and C.~D. Manning.
\newblock Probabilistic models of language processing and acquisition.
\newblock {\em Trends in Cognitive Sciences}, 10(7):335--344, 2006.

\bibitem{chater2006a}
N.~Chater, J.~B. Tenenbaum, and A.~Yuille.
\newblock Probabilistic models of cognition: Conceptual foundations.
\newblock {\em Trends in Cognitive Sciences}, 10(7):287--291, 2006.

\bibitem{dasgupta2016}
I.~Dasgupta, E.~Schulz, and S.~J. Gershman.
\newblock Where do hypotheses come from?
\newblock Technical report, Center for Brains, Minds and Machines (CBMM), 2016.

\bibitem{gao2006}
J.~Gao, V.~A. Billock, I.~Merk, W.~Tung, K.~D. White, J.~Harris, and V.~P.
  Roychowdhury.
\newblock Inertia and memory in ambiguous visual perception.
\newblock {\em Cognitive Processing}, 7(2):105--112, 2006.

\bibitem{gershman2012}
S.~J. Gershman, E.~Vul, and J.~B. Tenenbaum.
\newblock Multistability and perceptual inference.
\newblock {\em Neural Computation}, 24(1):1--24, 2012.

\bibitem{geyer1991}
C.~J. Geyer.
\newblock Markov chain monte carlo maximum likelihood.
\newblock 1991.

\bibitem{gilden1997}
D.~L. Gilden.
\newblock Fluctuations in the time required for elementary decisions.
\newblock {\em Psychological Science}, 8(4):296--301, 1997.

\bibitem{gilden1995}
D.~L. Gilden, T.~Thornton, and M.~W. Mallon.
\newblock $1/f$ noise in human cognition.
\newblock {\em Science}, 267(5205):1837, 1995.

\bibitem{gonzalez2008}
M.~C. Gonzalez, C.~A. Hidalgo, and A.-L. Barabasi.
\newblock Understanding individual human mobility patterns.
\newblock {\em Nature}, 453(7196):779--782, 2008.

\bibitem{griffiths2011}
T.~L. Griffiths and J.~B. Tenenbaum.
\newblock Predicting the future as bayesian inference: people combine prior
  knowledge with observations when estimating duration and extent.
\newblock {\em Journal of Experimental Psychology: General}, 140(4):725, 2011.

\bibitem{hastings1970}
W.~K. Hastings.
\newblock Monte carlo sampling methods using markov chains and their
  applications.
\newblock {\em Biometrika}, 57(1):97--109, 1970.

\bibitem{kello2010}
C.~T. Kello, G.~D. Brown, R.~Ferrer-i Cancho, J.~G. Holden,
  K.~Linkenkaer-Hansen, T.~Rhodes, and G.~C. Van~Orden.
\newblock Scaling laws in cognitive sciences.
\newblock {\em Trends in Cognitive Sciences}, 14(5):223--232, 2010.

\bibitem{kemp2009}
C.~Kemp and J.~B. Tenenbaum.
\newblock Structured statistical models of inductive reasoning.
\newblock {\em Psychological Review}, 116(1):20, 2009.

\bibitem{lieder2012}
F.~Lieder, T.~Griffiths, and N.~Goodman.
\newblock Burn-in, bias, and the rationality of anchoring.
\newblock In {\em Advances in Neural Information Processing Systems}, pages
  2690--2798, 2012.

\bibitem{mackay2003}
D.~J. MacKay.
\newblock {\em Information theory, inference and learning algorithms}.
\newblock Cambridge university press, 2003.

\bibitem{metropolis1953}
N.~Metropolis, A.~W. Rosenbluth, M.~N. Rosenbluth, A.~H. Teller, and E.~Teller.
\newblock Equation of state calculations by fast computing machines.
\newblock {\em The Journal of Chemical Physics}, 21(6):1087--1092, 1953.

\bibitem{ramos2004}
G.~Ramos-Fern{\'a}ndez, J.~L. Mateos, O.~Miramontes, G.~Cocho, H.~Larralde, and
  B.~Ayala-Orozco.
\newblock L{\'e}vy walk patterns in the foraging movements of spider monkeys
  (ateles geoffroyi).
\newblock {\em Behavioral Ecology and Sociobiology}, 55(3):223--230, 2004.

\bibitem{rhodes2011}
T.~Rhodes, C.~T. Kello, and B.~Kerster.
\newblock Distributional and temporal properties of eye movement trajectories
  in scene perception.
\newblock In {\em 33th Annual Meeting of the Cognitive Science Society}, 2011.

\bibitem{rhodes2007}
T.~Rhodes and M.~T. Turvey.
\newblock Human memory retrieval as l{\'e}vy foraging.
\newblock {\em Physica A: Statistical Mechanics and its Applications},
  385(1):255--260, 2007.

\bibitem{sanborn2016}
A.~N. Sanborn and N.~Chater.
\newblock Bayesian brains without probabilities.
\newblock {\em Trends in Cognitive Sciences}, 20(12):883--893, 2016.

\bibitem{sanborn2013}
A.~N. Sanborn, V.~K. Mansinghka, and T.~L. Griffiths.
\newblock Reconciling intuitive physics and newtonian mechanics for colliding
  objects.
\newblock {\em Psychological Review}, 120(2):411, 2013.

\bibitem{shlesinger1995}
M.~F. Shlesinger, G.~M. Zaslavsky, and U.~Frisch.
\newblock L{\'e}vy flights and related topics in physics.
\newblock {\em Lecture Notes in Physics}, 450:52, 1995.

\bibitem{sims2008}
D.~W. Sims, E.~J. Southall, N.~E. Humphries, G.~C. Hays, C.~J. Bradshaw, J.~W.
  Pitchford, A.~James, M.~Z. Ahmed, A.~S. Brierley, M.~A. Hindell, et~al.
\newblock Scaling laws of marine predator search behaviour.
\newblock {\em Nature}, 451(7182):1098--1102, 2008.

\bibitem{swendsen1986}
R.~H. Swendsen and J.-S. Wang.
\newblock Replica monte carlo simulation of spin-glasses.
\newblock {\em Physical Review Letters}, 57(21):2607, 1986.

\bibitem{thornton2005}
T.~L. Thornton and D.~L. Gilden.
\newblock Provenance of correlations in psychological data.
\newblock {\em Psychonomic Bulletin \& Review}, 12(3):409--441, 2005.

\bibitem{troyer1997}
A.~K. Troyer, M.~Moscovitch, and G.~Winocur.
\newblock Clustering and switching as two components of verbal fluency:
  evidence from younger and older healthy adults.
\newblock {\em Neuropsychology}, 11(1):138, 1997.

\bibitem{vanorden2003}
G.~C. Van~Orden, J.~G. Holden, and M.~T. Turvey.
\newblock Self-organization of cognitive performance.
\newblock {\em Journal of Experimental Psychology: General}, 132(3):331, 2003.

\bibitem{viswanathan1996}
G.~M. Viswanathan, V.~Afanasyev, S.~Buldyrev, E.~Murphy, et~al.
\newblock L{\'e}vy flight search patterns of wandering albatrosses.
\newblock {\em Nature}, 381(6581):413, 1996.

\bibitem{viswanathan1999}
G.~M. Viswanathan, S.~V. Buldyrev, S.~Havlin, M.~Da~Luz, E.~Raposo, and H.~E.
  Stanley.
\newblock Optimizing the success of random searches.
\newblock {\em Nature}, 401(6756):911--914, 1999.

\bibitem{vul2014}
E.~Vul, N.~Goodman, T.~L. Griffiths, and J.~B. Tenenbaum.
\newblock One and done? optimal decisions from very few samples.
\newblock {\em Cognitive science}, 38(4):599--637, 2014.

\bibitem{wagenmakers2004}
E.-J. Wagenmakers, S.~Farrell, and R.~Ratcliff.
\newblock Estimation and interpretation of $1/f^\alpha$ noise in human
  cognition.
\newblock {\em Psychonomic Bulletin \& Review}, 11(4):579--615, 2004.

\bibitem{ward2002}
L.~M. Ward.
\newblock {\em Dynamical Cognitive Science}.
\newblock MIT press, 2002.

\bibitem{wolpert2007}
D.~M. Wolpert.
\newblock Probabilistic models in human sensorimotor control.
\newblock {\em Human Movement Science}, 26(4):511--524, 2007.

\bibitem{xu2009}
J.~Xu and T.~L. Griffiths.
\newblock How memory biases affect information transmission: A rational
  analysis of serial reproduction.
\newblock In {\em Advances in Neural Information Processing Systems}, pages
  1809--1816, 2009.

\bibitem{yuille2006}
A.~Yuille and D.~Kersten.
\newblock Vision as bayesian inference: analysis by synthesis?
\newblock {\em Trends in Cognitive Sciences}, 10(7):301--308, 2006.

\end{thebibliography}

\end{document}